\pdfoutput=1
\documentclass{article} 
\usepackage{iclr2027_conference,times}

\usepackage{amsmath,amsfonts,bm}

\def\eqref#1{equation~\ref{#1}}

\def\1{\bm{1}}

\DeclareMathAlphabet{\mathsfit}{\encodingdefault}{\sfdefault}{m}{sl}
\SetMathAlphabet{\mathsfit}{bold}{\encodingdefault}{\sfdefault}{bx}{n}

\usepackage{hyperref}
\usepackage{url}
\usepackage{booktabs}
\usepackage{algorithm}
\usepackage{algorithmic}
\usepackage{subfigure}
\usepackage{soul}
\usepackage{color}
\usepackage{xcolor}
\usepackage{amsmath}
\usepackage{amsfonts}
\usepackage{multirow}
\usepackage{wrapfig}
\usepackage{graphicx}
\usepackage{xspace}
\usepackage{booktabs}
\usepackage{enumitem}
\usepackage{nameref}
\usepackage{array}
\usepackage{nicematrix} 
\usepackage{booktabs}
\usepackage{hyperref}
\usepackage[capitalize,noabbrev]{cleveref}
\usepackage[flushleft]{threeparttable}
\usepackage{multirow}
\usepackage{enumitem}
\usepackage{siunitx}
\usepackage{caption}
\usepackage{wrapfig}
\usepackage{titletoc}
\newcommand{\method}{\texttt{GRFBrain}\xspace }
\usepackage{caption}
\usepackage{booktabs,threeparttable}
\newcommand{\best}[1]{\mathbf{#1}}
\newcommand{\second}[1]{\underline{#1}}
\usepackage{newfloat}
\usepackage{listings}
\usepackage{pifont}
\usepackage{iclr2027_conference,times}

\newcommand{\cmark}{\textcolor{green!45!black}{\ding{51}}}
\newcommand{\xmark}{\textcolor{red!75!black}{\ding{55}}}
\newcommand{\na}{\textcolor{gray!70}{--}}

\usepackage[capitalize,noabbrev]{cleveref}
\title{GRFBrain: Graph-Structured Rectified Flows for EEG Dynamic Modeling}

\iclrfinalcopy
\author{
Haohui Jia\textsuperscript{$\spadesuit$},
\hspace{0.5mm}Zheng Chen\textsuperscript{$\clubsuit $},
\hspace{0.5mm}Jathurshan Pradeepkumar\textsuperscript{$\heartsuit$},
\hspace{0.5mm}\textbf{Xu Cao}\textsuperscript{$\diamondsuit$},
\hspace{0.5mm}
\textbf{Yasuko Matsubara}\textsuperscript{$\clubsuit $},
\\
\textbf{Yasushi Sakurai\textsuperscript{$\clubsuit $},
\hspace{0.5mm}Takashi Matsubara\textsuperscript{$\spadesuit$}}
\\
\hspace{0.5mm}\textsuperscript{$\spadesuit$} Faculty of Information Science and Technology, Hokkaido University, Japan\\
\hspace{0.5mm}\textsuperscript{$\clubsuit $}SANKEN, The University of Osaka, Japan\\
\hspace{0.5mm}\textsuperscript{$\heartsuit$}Department of Computer Science, University of Illinois Urbana-Champaign, USA\\
\hspace{0.5mm}\textsuperscript{$\diamondsuit$}PediaMed AI, USA\\
}

\begin{document}

\maketitle
\fancyhead[L]{Preprint}

\begin{abstract}
Forecasting time-varying functional connectivity from electroencephalography (EEG) requires modeling both history-dependent trends and structured variability across channels. Conditional flow matching provides a framework for distributional forecasting, yet it remains unclear whether graph-informed source distributions offer practical advantages over isotropic noise and strong deterministic predictors. We introduce a graph-structured residual flow framework that separates conditional mean prediction from stochastic residual transport. A history-only predictor estimates the future connectivity graph, while a graph Gaussian source encodes dependencies derived from past connectivity through a Laplacian-based covariance. A conditional velocity field transports source samples to future graph residuals, with transport time explicitly distinguished from physical EEG time. Our study identifies the conditions and controls needed to distinguish useful residual transport from improvements attributable to deterministic prediction, learned representations, and sampling effects. Our code is available at \url{https://github.com/HHJIAnmo/GRFBrain}.
\end{abstract}

\section{Introduction}
\label{sec:introduction}

Learning informative representations of electroencephalography (EEG) is central to clinical settings such as seizure detection~\citep{pradeepkumar2026tokenizing}. 
What makes this difficult is that the relevant information is not located in a single channel. 
A seizure, for instance, is characterized less by abnormal activity at one electrode than by the emergence, spread, and dissolution of synchrony across electrodes. 
Representations must therefore encode two coupled aspects of brain dynamics: how channel-wise activity evolves, and how the functional relationships among channels over time.

Predictive learning offers a route to EEG representation learning by requiring a model to infer future activity from its history. 
Existing graph-based methods pursue this idea through next-window prediction, evolving graph convolutions, or continuous-time spectral dynamics \citep{tang2022self,kotoge2025evobrain,jia2026odebrain}. 
However, these objectives are deterministic: regression maps each history to a single future, encouraging representations of the conditional mean while suppressing structured uncertainty.
This is particularly limiting near transitions, where similar histories may lead to different futures and discriminative cues can lie in deviations from the average trajectory.

\begin{table}[t]
\centering
\caption{
Comparison of representative methods for modeling EEG dynamics.
}
\label{tab:technical-comparison}
\setlength{\tabcolsep}{4.2pt}
\renewcommand{\arraystretch}{1.12}

\resizebox{\textwidth}{!}{
\begin{tabular}{llccc}
\toprule

\textbf{Model}
&
\textbf{Formulation}
&
\textbf{Continuous}
&
\textbf{Graphical}
&
\textbf{Stochastic}

\\
\midrule

CNN-LSTM
&
Spatial-temporal encoding
&
 \xmark & \xmark & \xmark 
\\

BIOT~\citep{yang2023biot}
&
Biosignal Transformer
&
\xmark & \xmark & \xmark 
\\

EvolveGCN~\citep{pareja2020evolvegcn}
&
Evolving graph convolution
&
\xmark & \cmark & \xmark 
\\

DCRNN
&
Diffusion-graph convolutional RNN
&
\xmark & \cmark & \xmark 
\\
\midrule
neural ODE
&
latent dynamics
&
 \cmark & \xmark  & \xmark 
\\

neural SDE
&
Stochastic latent dynamics
&
\cmark & \xmark  & \cmark 
\\

ODEBrain~\citep{jia2026odebrain}
&
Temporal-spatial latent dynamics
&
\cmark & \cmark  & \xmark 
\\


Flow Matching
&
Isotropic latent transport
&
\cmark & \xmark & \cmark 
\\


\textbf{\method}
&
\textbf{Conditioned joint graph flow}
&
 \cmark & \cmark & \cmark 
\\

\bottomrule
\end{tabular}
}

\begin{minipage}{0.99\textwidth}
\footnotesize
\cmark~denotes that the property is implemented;
\xmark~denotes that it is absent from the evaluated formulation.
\end{minipage}
\end{table}

Distributional objectives model multiple plausible futures rather than a single point estimate. Flow matching (FM) realizes this by learning a history-conditioned velocity field that transports a simple source distribution toward future states~\citep{lipman2023flow}, encouraging representations to capture both predictable trends and residual variability. In EEG, however, this variability is structured across channels and reflected in recent functional connectivity, whereas standard FM starts from isotropic Gaussian noise that treats channels and channel pairs as exchangeable, leaving the velocity field to recover these dependencies during transport. The central question is therefore not whether isotropic noise is sufficiently expressive, but whether a history-informed source defines a more informative predictive task for representation learning.

\begin{figure}[t]
    \centering\includegraphics[width=0.99\linewidth]{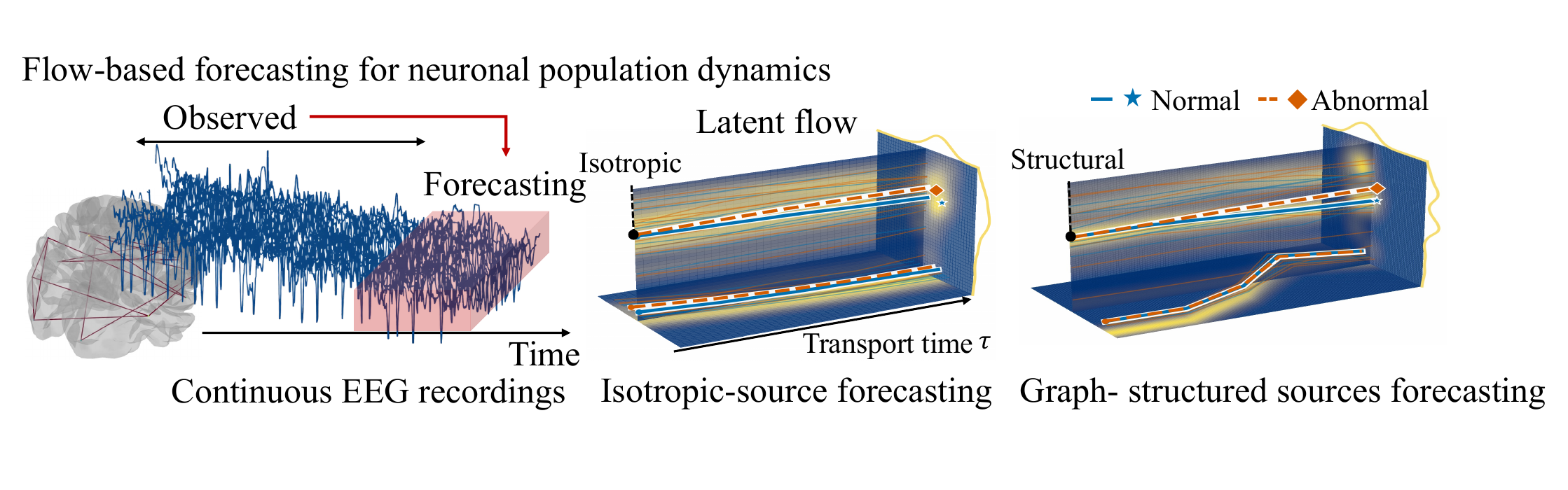}
   \caption{(Left) Continuous EEG real-time neuronal activity recordings. (Mid) An isotropic gaussian source-based method learns dynamic representations.
   (Right) Ours provides a graph conditioned source-based approach for learning the neuronal population dynamics.
   }
   \label{fig:figure1}
\end{figure}

This observation suggests that the source is not merely a sampling convenience but part of the learning task, and hence an inductive bias on the learned representation, as shown in Figure\@\ref{fig:figure1}.
We therefore ask: \emph{can the channel dependencies observed in the history define the stochastic prediction task itself, rather than serving only as conditioning input to the encoder?}

In this paper, we answer this question with \method, a graph-structured residual flow framework for predictive EEG representation learning.
\method first decomposes the future EEG state into a history-derived reference and a residual, allowing the flow objective to focus on the variation that remains beyond the predictable trend.
It then constructs a history-conditioned Gaussian source whose covariance is derived from the recent functional graph, so that source perturbations reflect the channel dependencies observed in each sample.
The source energy is normalized to match that of an IID Gaussian control, so that their comparison isolates the role of relational structure from that of noise magnitude.
Finally, a joint node--edge velocity network transports node and edge residuals in a shared state space, so that their bidirectional interactions shape the learned representation.
For downstream prediction, the velocity network is evaluated once at the history-derived reference state, which corresponds to zero in residual coordinates.
This readout requires neither source sampling nor numerical integration.
As summarized in Table~\ref{tab:technical-comparison}, existing methods are typically either graph-aware but deterministic, or stochastic without adaptive graph structure.
Our contributions are threefold:
\begin{itemize}[left=0pt]
    \item We formulate predictive EEG representation learning as conditional transport in a joint node--edge residual space, separating the history-predictable component of the future from its remaining structured variability.
    
    \item We introduce a history-conditioned graph Gaussian source with energy matched to an IID baseline, together with a coupled node--edge velocity network that models interactions between channel activity and functional connectivity.
    
    \item We evaluate GSRF on seizure detection and conduct controlled studies of source geometry, node--edge coupling, transport targets, and forecasting horizon. GSRF improves F1 over deterministic graph-dynamics and standard flow-matching baselines, while the ablations isolate the benefit of history-adaptive source structure.
\end{itemize}

\section{Related Works}

\noindent \textbf{Temporal graph methods for modeling EEG dynamics. }
Multichannel EEG can be represented as a graph, with channels as nodes and inter-channel relations as edges, motivating models that jointly capture temporal activity and spatial dependencies. Early work introduced temporal graph convolutions for seizure detection~\citep{covert2019tgcn}, followed by diffusion-convolutional recurrent networks with future-signal pretraining on geometric or correlation-based graphs~\citep{tang2022self}. Later methods extended this framework through state-space modeling and dynamic graph learning in GraphS4mer~\citep{tang2023graphssm}, efficient residual recurrent updates in REST~\citep{afzal2024rest}, and explicit time-varying node-edge states in EvoBrain~\citep{kotoge2025evobrain}. Continuous-time extensions include BrainODE for reconstructing irregularly sampled fMRI with graph-aided neural ODEs~\citep{han2024brainode} and ODEBrain for learning continuous latent EEG graph trajectories under future-graph supervision~\citep{jia2026odebrain}. Despite this progression from temporal encoding to dynamic and continuous trajectories, these methods produce discriminative representations, point forecasts, or single trajectories rather than conditional distributions over future node-edge states.

\noindent \textbf{Generative methods for brain modeling. }
Generative graph models learn joint distributions over node and edge variables. GDSS evolves continuous graph states through coupled stochastic differential equations, while DiGress denoises categorical graphs in discrete spaces \citep{jo2022score,vignac2023digress}; both target graph synthesis rather than EEG-conditioned forecasting. Flow Matching (FM) learns continuous transport along prescribed probability paths \citep{lipman2023flow}, with extensions addressing graph generation and geometry \citep{eijkelboom2024variational,jiang2026bwflow}, source--target coupling and conditional sources \citep{tong2024minibatchot,chen2025carflow,issachar2025conditionalprior,kim2026bettersource}, and topology-aware priors or objectives \citep{borovitskiy2021matern,wyrwal2026topological}. Closest to our setting, GiFlow uses graph-informed priors for spatiotemporal imputation but does not jointly transport evolving edges \citep{zhang2026giflow}, whereas DiffeoCFM models brain-connectivity distributions without conditioning their evolution on EEG history \citep{collas2025diffeocfm}. \method addresses this gap through a history-conditioned graph Gaussian source, reference-centered future residuals, and a shared velocity field for joint node-edge transport.

\section{Preliminary and Problem Formulation}
\label{sec:problem}

\noindent\textbf{Conditional Flow Matching.}
Given condition $\mathcal{C}$, we sample
$\boldsymbol{z}_0\sim p_0(\cdot\mid\mathcal{C})$,
$\boldsymbol{z}_1\sim p_1(\cdot\mid\mathcal{C})$, and
$\tau\sim\mathcal{U}[0,1]$. The linear path
$\boldsymbol{z}_\tau=(1-\tau)\boldsymbol{z}_0+\tau\boldsymbol{z}_1$
has target velocity
$\boldsymbol{u}_\tau=\boldsymbol{z}_1-\boldsymbol{z}_0$. Conditional Flow
Matching (CFM) learns
\begin{equation}
\mathcal{L}_{\mathrm{CFM}}(\theta)
=
\mathbb{E}_{\mathcal{C},\boldsymbol{z}_0,\boldsymbol{z}_1,\tau}
\left[
\left\|
v_\theta(\boldsymbol{z}_\tau,\tau\mid\mathcal{C})
-\boldsymbol{u}_\tau
\right\|_2^2
\right].
\label{eq:prelim-cfm-objective}
\end{equation}
Although $\boldsymbol{u}_\tau$ is defined per sampled pair, the optimal
regressor is the marginal velocity
\begin{equation}
v^\star(\boldsymbol{z},\tau\mid\mathcal{C})
=
\mathbb{E}\!\left[
\boldsymbol{z}_1-\boldsymbol{z}_0
\mid
\boldsymbol{z}_\tau=\boldsymbol{z},\mathcal{C}
\right],
\label{eq:prelim-cfm-minimizer}
\end{equation}
whose flow transports $p_0(\cdot\mid\mathcal{C})$ to
$p_1(\cdot\mid\mathcal{C})$ \citep{lipman2023flow}. Training requires no
numerical integration, while generation solves
$d\boldsymbol{z}_\tau/d\tau
=v_\theta(\boldsymbol{z}_\tau,\tau\mid\mathcal{C})$.
Here, $\tau$ is an artificial transport coordinate, distinct from the physical EEG window time $s$; the learned velocity therefore describes distributional transport rather than physiological dynamics.

\noindent\textbf{Source-Target Mismatch in EEG Graph Transport.}
Standard CFM uses the isotropic source
$p_0(\boldsymbol{z}\mid\mathcal{C})
=\mathcal{N}(\boldsymbol{0},\boldsymbol{I})$, as an independent identically distributed (IID).
Although expressive in principle, this source ignores the relational geometry
available in the past EEGs. Future EEG graphs couple channel-wise node
features with pairwise edges, and their unpredictable component retains history-dependent cross-channel structure. An isotropic source needs a finite-capacity velocity field to learn the conditional shift, and node-edge dependence. This mismatch raises two questions.

\noindent\textbf{1. How should the state and source be defined?}
The future state should be centered by its temporal component, with
source and target defined in the same residual node-edge space. The source should use observed history, encode the graph topology to control the total variance, so that a comparison with an IID source isolates the covariance structure rather than the noise scale.

\noindent\textbf{2. How should nodes and edges be transported jointly?}
Because node activity and functional connectivity are coupled, the velocity
field should support bidirectional node-to-edge and edge-to-node interactions
instead of fixing the adjacency matrix or learning independent fields.

These requirements motivate our predictor-centered residual state,
temporal-conditioned graph source, and shared node-edge velocity field.
\section{Methodology}
\label{sec:methodology}
\begin{figure*}[t]
\centering
\includegraphics[width = 0.99\textwidth]{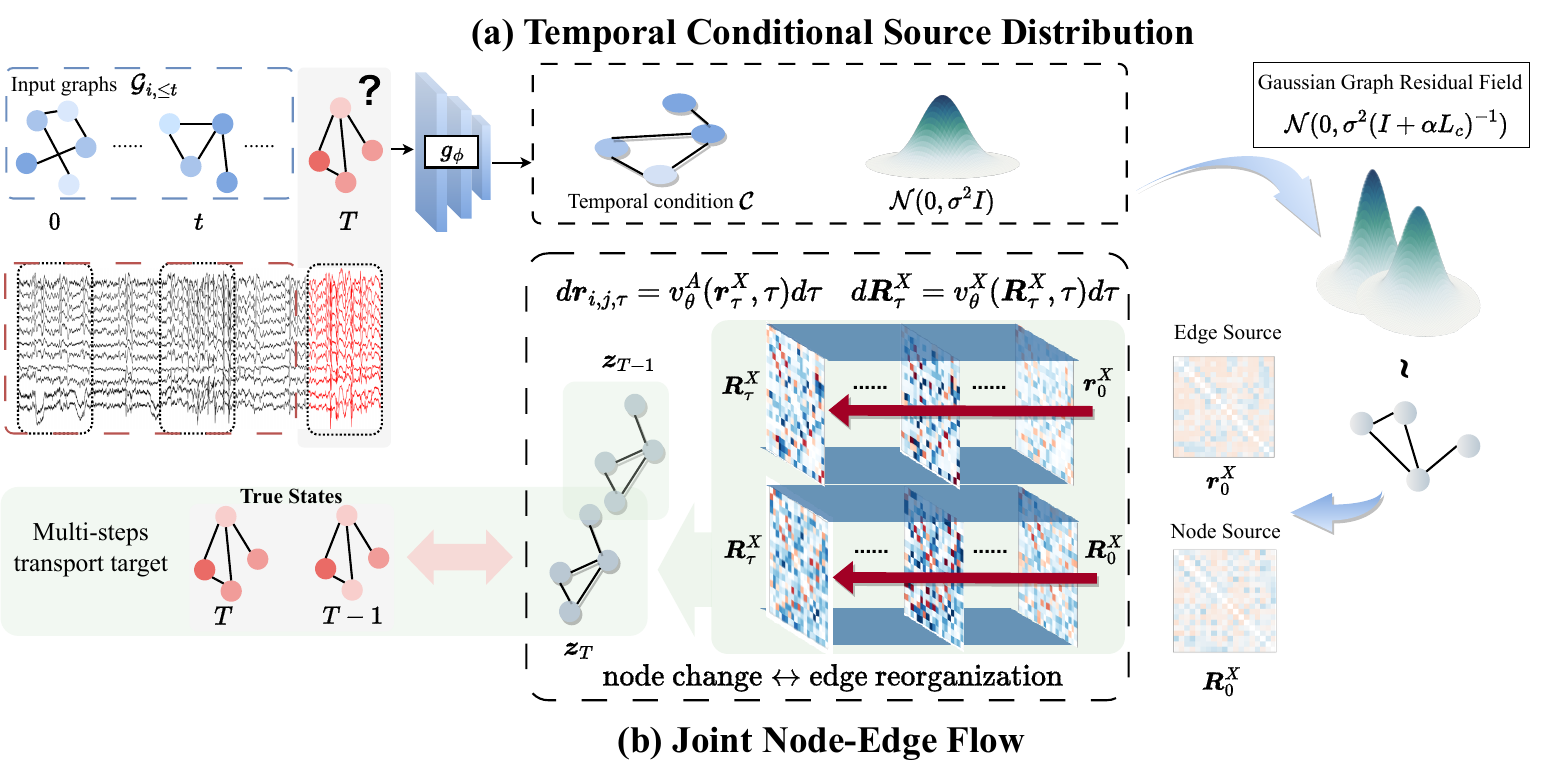}
\caption{
Overview of the proposed \method.
(a) Temporal EEG features and functional association graphs define
state-dependent states and a graph Gaussian residual source,
whose covariance is adapted to the observed graph states.
(b) A joint node-edge velocity network transports node and edge
residuals along a shared flow coordinate, coupling channel-wise state variations.
}
\label{fig: proposal}
\end{figure*}

Figure \ref{fig: proposal} shows the system overview of \method. First, it decomposes the temporal EEG graph embedding with sequential encoding as condition states to construct a
graph-structured Gaussian source. Second, a Joint Node-Edge velocity field between source and target state. 

\label{sec: method}

\subsection{History-Conditioned Graph Gaussian Source}
\label{sec:source}

Let
\(\mathcal C=\{(\boldsymbol X_s,\boldsymbol A_s)\}_{s=t-H+1}^{t}\)
denote an EEG history of \(H\) physical windows, where
\(\boldsymbol X_s\in\mathbb R^{N\times F}\) contains channel-wise spectral
features and \(\boldsymbol A_s\in\mathbb R^{N\times N}\) is a directed
top-\(k\) graph. We represent its undirected continuous edge state by
\(\boldsymbol a_s=\mathcal U(\boldsymbol S_s)\in\mathbb R^m\), where
\(\boldsymbol S_s=(\boldsymbol A_s+\boldsymbol A_s^\top)/2\) has zero
diagonal, \(\mathcal U\) extracts its strict upper triangle, and
\(m=N(N-1)/2\).

We model the future graph relative to a history-dependent reference. Let
\(\boldsymbol M_X\in\mathbb R^{N\times F}\) be the mean historical node state
and \(\boldsymbol m_A\in\mathbb R^m\) the edge reference produced by a frozen
graph forecaster \(g_\phi\). The target residuals are
\begin{equation}
\boldsymbol M_X
=\frac{1}{H}\sum_{s=t-H+1}^{t}\boldsymbol X_s,
\qquad
\boldsymbol R_1^X
=\boldsymbol X_{t+1}-\boldsymbol M_X,
\qquad
\boldsymbol r_1^A
=\boldsymbol a_{t+1}-\boldsymbol m_A.
\label{eq:conditional-residuals}
\end{equation}
This decomposition assigns the predictable component to the reference model
and leaves the conditional variation to the flow. Unlike an independent and identically distributed (i.i.d.) Gaussian source,
our source uses the relational geometry of the observed history. Define
\(\overline{\boldsymbol S}_{\mathcal C}
=H^{-1}\sum_{s=t-H+1}^{t}\boldsymbol S_s\) and construct
\begin{equation}
\boldsymbol L_{\mathcal C}
=
\boldsymbol I_N
-\boldsymbol D_{\mathcal C}^{-1/2}
 \overline{\boldsymbol S}_{\mathcal C}
 \boldsymbol D_{\mathcal C}^{-1/2},
\qquad
\boldsymbol K_{\mathcal C}
=
\frac{
N(\boldsymbol I_N+\alpha\boldsymbol L_{\mathcal C})^{-\nu}
}{
\operatorname{tr}
[(\boldsymbol I_N+\alpha\boldsymbol L_{\mathcal C})^{-\nu}]
},
\qquad
\boldsymbol G_{\mathcal C}
=\boldsymbol K_{\mathcal C}^{1/2}.
\label{eq:ggrf-operator}
\end{equation}
Here, \(\boldsymbol D_{\mathcal C}\) is the stabilized degree matrix,
\(\alpha\geq0\), and \(\nu>0\). Trace normalization gives
\(\operatorname{tr}(\boldsymbol K_{\mathcal C})=N\), while
\((1+\alpha\lambda_i)^{-\nu}\) suppresses high graph-frequency modes. Node and edge residuals are sampled as
\begin{equation}
\boldsymbol R_0^X
=
\sigma_X\boldsymbol G_{\mathcal C}\boldsymbol\epsilon_X,
\qquad
\boldsymbol r_0^A
=
\sigma_A s_{\mathcal C}\,
\mathcal U\!\left(
\boldsymbol G_{\mathcal C}
\boldsymbol\Xi_A
\boldsymbol G_{\mathcal C}
\right),
\qquad
\boldsymbol\Xi_A=\mathcal U^{-1}(\boldsymbol\epsilon_A),
\label{eq:joint-ggrf-source}
\end{equation}
where \(\boldsymbol\epsilon_X\) and \(\boldsymbol\epsilon_A\) are independent
standard Gaussian innovations, and \(s_{\mathcal C}\) normalizes the edge
energy. The two sources are conditionally independent but share
\(\boldsymbol G_{\mathcal C}\), so their covariance follows the historical
graph while their total energies remain matched to the i.i.d.\ control. In
the original coordinates, the source is centered at
\((\boldsymbol M_X,\boldsymbol m_A)\). The graph is used only as an empirical
conditioning structure and is not interpreted as an anatomical or causal
brain network. Appendix~\ref{app:conditioned-ggrf} provides the forecaster
objective, numerical stabilization, and covariance derivations.

\subsection{Joint Node--Edge Residual Flow}
\label{sec:velocity}

We model node and edge residuals with a shared velocity field
\[
v_\theta:
\mathbb R^{N\times F}\times\mathbb R^m\times[0,1]\times\mathcal C
\rightarrow
\mathbb R^{N\times F}\times\mathbb R^m,
\]
implemented by a joint node--edge Transformer
\citep{jo2022score,vignac2023digress}. It maintains one node token per
channel and one symmetric edge token per unordered channel pair, including
pairs with zero observed weight.

Let \(\boldsymbol c_i\) denote the temporal encoding of channel \(i\), and let
\(\boldsymbol c_A\) summarize the edge history. The global condition
\(\boldsymbol q_\tau
=[N^{-1}\sum_i\boldsymbol c_i;\boldsymbol c_A;
\operatorname{Emb}(\tau)]\)
modulates each block through feature-wise linear modulation (FiLM)
\citep{perez2018film}. Initial tokens are
\begin{equation}
\begin{aligned}
\boldsymbol h_i^{(0)}
&=
P_X([\boldsymbol R_{\tau,i}^X;\boldsymbol M_{X,i}])
+P_N(\boldsymbol c_i),
\\
\boldsymbol e_{ij}^{(0)}
&=
P_A([r_{\tau,ij}^A;m_{A,ij}])
+P_P(\chi(\boldsymbol c_i,\boldsymbol c_j))
+P_O([L_{\mathcal C,ij};K_{\mathcal C,ij}]),
\qquad i<j,
\end{aligned}
\label{eq:initial-tokens}
\end{equation}
where \(P_X,P_N,P_A,P_P,P_O\) are learned projections and
\(\chi(\boldsymbol u,\boldsymbol v)
=[\boldsymbol u+\boldsymbol v;
|\boldsymbol u-\boldsymbol v|;
\boldsymbol u\odot\boldsymbol v]\)
is symmetric in its arguments.

Each interaction block uses edge tokens as attention biases and incident-edge
messages for node updates. The updated endpoint tokens then revise their edge
through the same symmetric descriptor:
\begin{equation}
\begin{aligned}
\boldsymbol h_i^{(\ell+1)}
&=
\operatorname{NodeBlock}_{\ell}
\left(
\boldsymbol h_i^{(\ell)},
\{\boldsymbol e_{ij}^{(\ell)}\}_{j};
\boldsymbol q_\tau
\right),
\\
\boldsymbol e_{ij}^{(\ell+1)}
&=
\operatorname{EdgeBlock}_{\ell}
\left(
\boldsymbol e_{ij}^{(\ell)},
\chi(\boldsymbol h_i^{(\ell+1)},
     \boldsymbol h_j^{(\ell+1)});
\boldsymbol q_\tau
\right).
\end{aligned}
\label{eq:node-edge-interaction}
\end{equation}
After \(J\) blocks, separate affine heads produce
\(\boldsymbol v_\theta^X\in\mathbb R^{N\times F}\) and
\(\boldsymbol v_\theta^A\in\mathbb R^m\). Thus, edges guide node updates while
node states jointly determine edge updates.

\subsection{Conditional Flow-Matching Objective}
\label{sec:objectives}

For each future target, we sample
\((\boldsymbol R_0^X,\boldsymbol r_0^A)\) from the conditional source and draw
a shared \(\tau\sim\mathcal U(0,1)\). Node and edge residuals follow
\begin{equation}
\begin{aligned}
\boldsymbol R_\tau^X
&=(1-\tau)\boldsymbol R_0^X+\tau\boldsymbol R_1^X,
&
\boldsymbol u^X
&=\boldsymbol R_1^X-\boldsymbol R_0^X,
\\
\boldsymbol r_\tau^A
&=(1-\tau)\boldsymbol r_0^A+\tau\boldsymbol r_1^A,
&
\boldsymbol u^A
&=\boldsymbol r_1^A-\boldsymbol r_0^A.
\end{aligned}
\label{eq:flow-path}
\end{equation}
Source and target residuals are paired independently given \(\mathcal C\);
hence, this is conditional linear Flow Matching rather than an
optimal-transport coupling.

To balance the two state spaces, let \(q_X\) and \(q_A\) be the fixed
per-coordinate target-velocity energies estimated from the training set:
\(q_X=\max\{
\widehat{\mathbb E}_{\mathrm{train}}
[\|\boldsymbol u^X\|_F^2/(NF)],\epsilon_q\}\) and
\(q_A=\max\{
\widehat{\mathbb E}_{\mathrm{train}}
[\|\boldsymbol u^A\|_2^2/m],\epsilon_q\}\).
The normalized conditional Flow Matching (CFM) objective is
\begin{equation}
\mathcal L_{\mathrm{CFM}}(\theta)
=
\mathbb E\!\left[
\frac{
\|\boldsymbol v_\theta^X-\boldsymbol u^X\|_F^2
}{NFq_X}
+
\frac{
\|\boldsymbol v_\theta^A-\boldsymbol u^A\|_2^2
}{mq_A}
\right],
\label{eq:fm-loss}
\end{equation}
where the expectation covers training histories and targets, conditional
source samples, and \(\tau\). The objective remains standard CFM
\citep{lipman2023flow}; the proposed design changes the residual target,
source covariance, and joint node--edge state space.

\section{Experiments}
In this section, we conduct experiments to answer the following research questions:  \\
- RQ1.
Does \method strengthen seizure detection capacity through temporal-conditioned flow matching?\\
- RQ2.
How does source geometry affect the development of latent transport? \\
- RQ3.
Does the source-target objective of $\mathcal L_{\mathrm{CFM}}$ facilitate dynamic optimization?\\
More detailed experiment settings can be found in the Appendix \ref{apdx:setting}.

\subsection{Experimental Setup}

\noindent \textbf{Tasks.} \quad We evaluate \method on two downstream tasks: \textit{seizure detection} and \textit{abnormal EEG classification}. These complementary tasks assess the utility of the learned EEG representations in identifying seizure-related activity and distinguishing abnormal from EEG patterns. Detailed descriptions of the datasets are provided in Section~\ref{sec:datasets}.

\begin{wraptable}{r}{0.5\linewidth}  
  \vspace{-12pt}
  \centering
  \caption{Results (AUROC$\uparrow$, F1$\uparrow$) on TUSZ (12s and 60s seizure detection).
  The upper block compares \method with discrete and continuous baselines (bold = best);
  the lower block ablates the node-edge interaction routes within \method.}
  \label{tab:RQ2-2}
  \small
  \setlength{\tabcolsep}{2pt}
  \begin{tabular}{c l c c c}
    \toprule
    \multirow{2}{*}{Model} & Method & T(s) & AUROC & F1 \\
    \cmidrule(lr){2-5}
    \multirow{4}{*}{\rotatebox[origin=c]{90}{Flow-based}}
      & BIOT    & 12  &  0.772$\pm$0.006 &  0.294$\pm$0.006\\
      &             & 60 & 0.642$\pm$0.009 & 0.256$\pm$0.003\\
      \cmidrule(lr){2-5}
      & DCRNN     & 12  & 0.825$\pm$0.002 & 0.416$\pm$0.009\\
      &             & 60 & 0.802$\pm$0.003 & 0.413$\pm$0.005\\
      \cmidrule(lr){2-5}
      & Flow matching     & 12  & 0.813$\pm$0.002 & 0.372$\pm$0.013\\
      &             & 60 & 0.725$\pm$0.006 & 0.311$\pm$0.028\\
      \cmidrule(lr){2-5}
      & \method     & 12  & \bfseries 0.877$\pm$0.003 & \bfseries 0.523$\pm$0.014\\
      &             & 60 & \bfseries 0.831$\pm$0.004 & \bfseries 0.488$\pm$0.032\\
    \midrule
    \multirow{4}{*}{\rotatebox[origin=c]{90}{\method}}
      & Edge-to-Node    & 12  &  0.867$\pm$0.004 &  0.488$\pm$0.007\\
      &           & 60 &  0.821$\pm$0.034 &  0.424$\pm$0.003\\
      \cmidrule(lr){2-5}
      & Node-to-Edge     & 12  & 0.848$\pm$0.017 & 0.462$\pm$0.013\\
      &             & 60 & 0.817$\pm$0.029 & 0.414$\pm$0.047\\
      \cmidrule(lr){2-5}
     & No-cross      & 12 & 0.673$\pm$0.007 & 0.374$\pm$0.033\\
      &             & 60 & 0.519$\pm$0.006 & 0.334$\pm$0.017\\
    \bottomrule
  \end{tabular}
\end{wraptable}

\noindent \textbf{Baseline methods.} We organize the baselines with the following learning paradigms.
First, we consider conventional sequence-based EEG models, including
CNN-LSTM~\citep{9175641} and the Transformer-based biosignal model
BIOT~\citep{yang2023biot}.
Second, we compare with discrete time graph dynamics models,
including DCRNN~\citep{li2017diffusion} and
EvolveGCN~\citep{pareja2020evolvegcn}, which model temporal graph
evolution through recurrent mechanisms.
We also include continuous time dynamics models.
ODE-RNN~\citep{rubanova2019latent}, a neural stochastic differential equation (SDE) model~\citep{liu2019neural},
graph differential equations (GDEs)~\citep{poli2019graph},
and ODEBrain~\citep{jia2026odebrain}, which explicitly models
continuous EEG graph dynamics.
Finally, we also compare with flow-based predictive baselines, including standard Flow matching~\citep{lipman2023flow} and Rectified flow~\citep{liu2022flow}. 

\begin{table*}[t]
\centering
\small
\setlength{\tabcolsep}{5pt}
\caption{Main results on TUSZ (12s seizure detection) and TUAB abnormal EEG classification,
reported as mean $\pm$ standard deviation over three seeds. \textbf{Bold} and \underline{underline} indicate the best and second-best
results for each metric, respectively.
$\dagger$ denotes models trained with a multi-step predictive objective,
and $\ddagger$ denotes models trained with a single-step predictive objective.
}

\label{tab:main_tusz_tuab}
\begin{threeparttable}
\resizebox{0.99\linewidth}{!}{
\begin{tabular}{lcccccc}
\toprule
\multirow{2}{*}{Method} 
& \multicolumn{3}{c}{\textbf{TUSZ}} & \multicolumn{3}{c}{\textbf{TUAB}} \\
\cmidrule(lr){2-4}\cmidrule(lr){5-7}
& {Acc} & {F1} & {AUROC}
& {Acc} & {F1} & {AUROC} \\
\midrule
CNN-LSTM      & $0.735 \pm 0.003$ & $0.347 \pm 0.012$ & $0.757 \pm 0.003$ & $0.741 \pm 0.002$ & $0.736 \pm 0.007$ & $0.813 \pm 0.003$ \\
BIOT          & $0.702 \pm 0.003$ & $0.294 \pm 0.006$ & $0.772 \pm 0.006$ & $0.717 \pm 0.002$ & $0.713 \pm 0.004$ & $0.788 \pm 0.002$ \\
EvolveGCN     & $0.769 \pm 0.002$ & $0.385 \pm 0.005$ & $0.791 \pm 0.004$ & $0.708 \pm 0.003$ & $0.707 \pm 0.002$ & $0.777 \pm 0.003$ \\
DCRNN         & $0.816 \pm 0.002$ & $0.416 \pm 0.009$ & $0.825 \pm 0.002$ & $0.768 \pm 0.004$ & $0.769 \pm 0.002$ & $0.848 \pm 0.002$ \\
\midrule
latent-ODE        & $0.827 \pm 0.004$ & $0.470 \pm 0.005$ & $0.849 \pm 0.004$ & $0.749 \pm 0.003$ & $0.745 \pm 0.002$ & $0.829 \pm 0.004$ \\
latent-ODE (RK4)  & $0.821 \pm 0.003$ & $0.465 \pm 0.001$ & $0.845 \pm 0.004$ & $0.746 \pm 0.002$ & $0.739 \pm 0.002$ & $0.823 \pm 0.003$ \\
ODE-RNN           & $0.802 \pm 0.002$ & $0.455 \pm 0.007$ & $0.855 \pm 0.003$ & $0.751 \pm 0.003$ & $0.744 \pm 0.004$ & $0.838 \pm 0.005$ \\
neural SDE        & $0.857 \pm 0.002$ & $0.467 \pm 0.003$ & $0.851 \pm 0.002$ & $0.768 \pm 0.003$ & $0.751 \pm 0.003$ & $0.834 \pm 0.002$ \\
GDEs              & $0.849 \pm 0.003$ & $0.475 \pm 0.005$ & $0.841 \pm 0.003$ & $0.757 \pm 0.003$ & $0.737 \pm 0.006$ & $0.823 \pm 0.004$ \\
Flow matching     & $0.874 \pm 0.002$ & $0.372 \pm 0.013$ & $0.813 \pm 0.002$ &$0.741 \pm 0.011$ & $0.719 \pm 0.007$ & $0.803 \pm 0.006$ \\
Rectified flow    & $0.721 \pm 0.001$ & $0.195 \pm 0.008$ & $0.566 \pm 0.002$ & $0.753 \pm 0.004$ & $0.721 \pm 0.004$ & $0.811 \pm 0.003$ \\
$\text{ODEBrain}^\dag$  & $0.869 \pm 0.003$ & $0.488 \pm 0.015$ & $0.875 \pm 0.005$ & $0.771 \pm 0.005$ & $0.770 \pm 0.005$ & $0.849 \pm 0.003$ \\
$\text{ODEBrain}^\ddag$ & $0.877 \pm 0.004$ & $0.496 \pm 0.017$ & $\best{0.881 \pm 0.006}$ & $0.778 \pm 0.003$ & $0.774 \pm 0.005$ & $0.857 \pm 0.005$ \\
\midrule
$\method^\dag$  & $\best{0.883 \pm 0.004}$ & $\second{0.503 \pm 0.012}$ & $0.876 \pm 0.001$ & $\best{0.789 \pm 0.006}$ & $\best{0.786 \pm 0.004}$ & $\best{0.878 \pm 0.004}$ \\

$\method^\ddag$ & $\second{0.878 \pm 0.001}$ & $\best{0.523 \pm 0.014}$ & $\second{0.877 \pm 0.003}$ & $\second{0.787 \pm 0.004}$ & $\second{0.783 \pm 0.004}$ & $\second{0.877 \pm 0.006}$ \\
\bottomrule
\end{tabular}
}
\begin{tablenotes}[flushleft]
\footnotesize
\item \na~denotes a configuration that has not been evaluated on the corresponding corpus.
\end{tablenotes}
\end{threeparttable}
\end{table*}

\begin{figure*}[htbp]
\centering
\includegraphics[width = 0.99\linewidth]{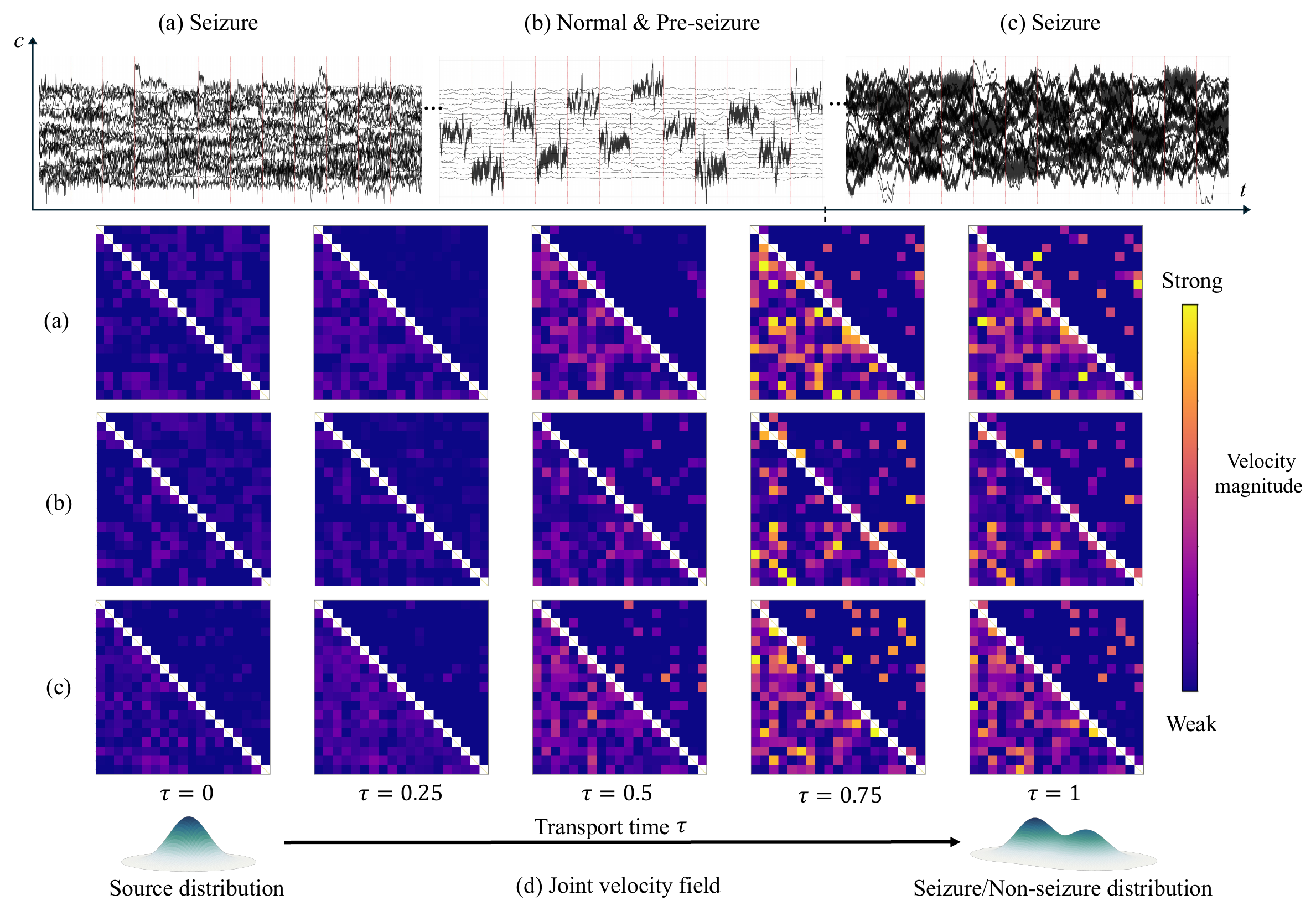}
\caption{Visualization of the proposed joint node-edge velocity field $v_{\theta}$ (middle) obtained by \method.
The learned velocity exists the distinct patterns of graph reconfiguration, which seizure represents higher and dense magnitude and non-seizure represents lower and sparse magnitude.
}
\label{fig:dynamic_field}
\end{figure*}

\noindent \textbf{Metrics.}
To answer \textbf{RQ1}, we evaluate the model using the Area Under the Receiver Operating Characteristic Curve (AUROC) and the F1 score. 
The AUROC measures the ability of the models across varying thresholds, while the F1 score highlights the balance between precision and recall at its optimal threshold for classification.
For \textbf{RQ2}, we measure the structural similarity of the predicted graph using the Global Jaccard Index (GJI) $
    \mathtt{GJI}(\mathcal{E}_{true},\mathcal{E}_{Pred}) = \frac{\lvert \mathcal{E}_{true}\cap \mathcal{E}_{Pred}\rvert}{\lvert \mathcal{E}_{true}\cup \mathcal{E}_{Pred}\rvert}$ \citep{castrillo2018dynamic}.
For \textbf{RQ3}, We compute the cosine similarity of predicted node embeddings.

\subsection{Results}

\subsubsection{Main Result} 

\textbf{RQ1} concerns the forecasting-based transport capability on EEG. Table~\ref{tab:main_tusz_tuab} compares \method with conventional EEG models, discrete graph-dynamics methods, continuous-time models, and generic flow-based predictors on TUSZ and TUAB. On TUSZ, the multi-step variant achieves the highest Accuracy of \(0.883\pm0.004\), improving over the corresponding ODEBrain variant across all three metrics. The single-step variant obtains the best F1 score of \(0.523\pm0.014\), an absolute gain of \(2.7\) percentage points over ODEBrain (\(0.496\pm0.017\)), while remaining competitive in AUROC (\(0.877\pm0.003\) versus \(0.881\pm0.006\)). On TUAB, both variants consistently outperform their ODEBrain counterparts, with the multi-step variant achieving the best Accuracy, F1, and AUROC of \(0.789\pm0.006\), \(0.786\pm0.004\), and \(0.878\pm0.004\), respectively.
Generic flow learning alone does not reproduce these gains. On TUSZ, standard flow matching attains a relatively high Accuracy of \(0.874\pm0.002\), but
its F1 and AUROC decrease to \(0.372\pm0.013\) and \(0.813\pm0.006\);
Rectified Flow performs worse across all three metrics. Taken together, these
results answer RQ1: distributional forecasting improves EEG representation
when flow matching is coupled with history-conditioned graph structure and
joint node-edge transport, rather than applied as a generic predictive
objective.

Beyond downstream performance, we further investigate whether the learned
joint field $v_{\theta}$ captures state-dependent reorganization of channel
associations. We analyze the edge-velocity field
$v_\theta^A$ across the transport time $\tau$.
Positive and negative velocities respectively indicate tendencies to
strengthen and weaken pairwise associations.
As shown in Figure\@\ref{fig:dynamic_field}, edge-velocity magnitudes are relatively weak near the source side and become increasingly structured toward larger $\tau$. 
Rather than exhibiting a uniform global change, the field
concentrates on subsets of channel pairs, suggesting that the learned representation is associated with selective graph reconfiguration.
\captionsetup{skip=2pt}
\begin{wrapfigure}[32]{r}{0.45\textwidth}
\centering
\includegraphics[width=0.99\linewidth]{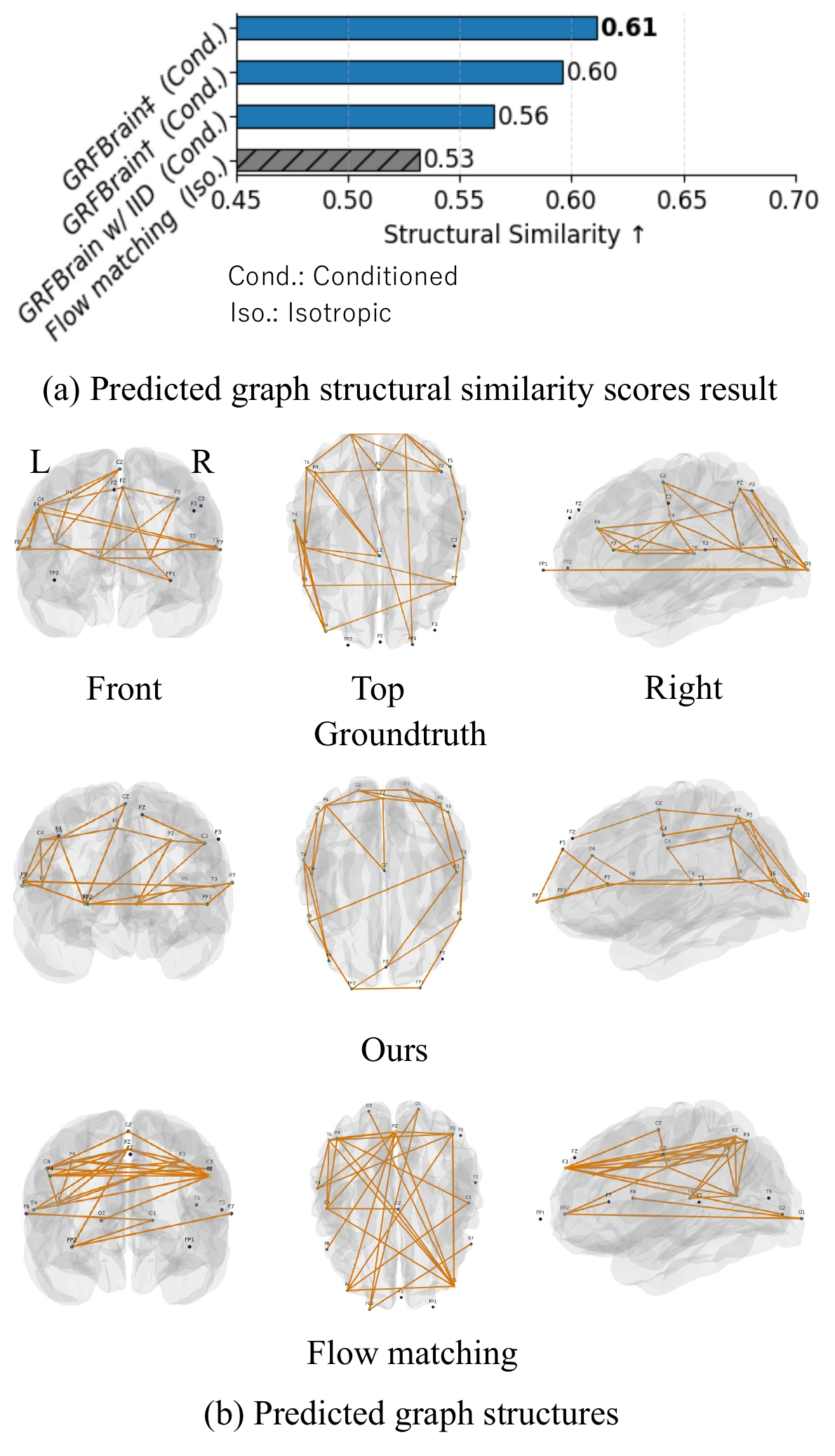}
\caption{Results on (a) graph similarity and (b) functional connections.}
\label{fig:brain_adj}
\vspace{-3.0cm}
\end{wrapfigure}
These results suggest that the joint velocity encourages representations that encode not only channel-wise spectral states but also how their
relational organization may be reconfigured. This provides a
complementary explanation for the downstream gains by interpreting the learned flow as an estimation of latent EEG dynamics.
Taken together, the quantitative and qualitative results answer \textbf{RQ1} affirmatively. \method improves seizure detection by coupling conditional flow matching with joint node-edge transport, thereby transforming temporal-dependent graph reconfiguration into class representations.

\begin{figure*}[t]
\centering
\includegraphics[width = 0.99\linewidth]{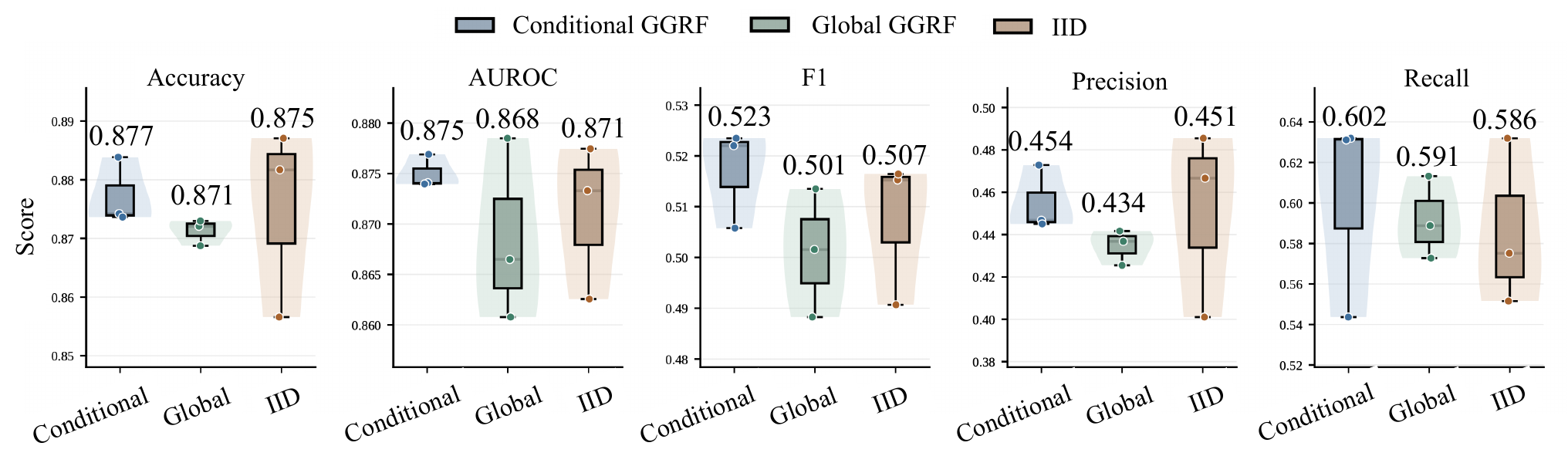}
\caption{Effect of source geometry in \method.
IID Gaussian, global GGRF, and our conditioned GGRF sources are compared across downstream classification metrics. Our conditional GGRF consistently improves the learned representation quality.
}
\label{fig:noise_bar}
\end{figure*}

\textbf{RQ2} concerns the geometry of noise.
Figure\@\ref{fig:noise_bar} compares the conditional GGRF with a global
GGRF and an IID Gaussian source. The conditional GGRF performs best across
all five metrics, achieving an Accuracy of $0.877$, an AUROC of $0.875$,
and an F1 score of $0.523$. Compared with the global GGRF, it improves F1
and AUROC by $2.2$ and $0.7$, respectively. It also outperforms the IID Gaussian in F1, AUROC, and Recall despite their similar Accuracy and Precision. The weaker performance of the global GGRF indicates that graph correlation alone is insufficient. The IID Gaussian discards inter-channel dependencies, whereas the global GGRF imposes the same relational geometry on each sample.
In contrast, the conditional GGRF adapts its covariance to the observed EEG history. Because the source determines both the interpolated states and their velocity targets, this sample-specific covariance produces transport paths aligned with the current graph structure. These results answer \textbf{RQ2}: latent
transport benefits from a history-conditioned source geometry rather than an isotropic or globally shared source, yielding more discriminative node-edge representations.

\noindent\textbf{Necessity of bidirectional node-edge interaction.}
Table~\ref{tab:RQ2-2} compares velocity-field designs on $12$\,s and
$60$\,s EEG clips. The full bidirectional field achieves the highest
AUROC/F1 of $0.877/0.523$ and $0.831/0.488$, respectively. Its advantage over standard FM shows that the generic objective alone is
insufficient, while the within-\method ablations isolate the interaction mechanism. Removing cross-stream messages reduces AUROC/F1 to $0.673/0.374$ and $0.519/0.334$, despite retaining both node and edge states. One-way interaction partially recovers performance, with Edge-to-Node consistently outperforming Node-to-Edge, likely because it directly updates the node representations consumed by the classifier. Neither direction alone matches bidirectional interaction. Relative to Edge-to-Node, the full field improves F1 by $0.035$ and $0.064$ at $12$\,s and $60$\,s, with an AUROC gain of $0.010$ in both settings. Thus, within this architecture, representing both state types is insufficient; reciprocal node-edge communication is required for the strongest downstream performance.

\noindent \textbf{RQ3} concerns consistency in the graphs with source-target objective of $\mathcal L_{\mathrm{CFM}}$.
Figure\@\ref{fig:brain_adj} compares the predicted functional graphs with their target structures. \method increases the graph-similarity score from $0.53$ for the FM to $0.61$. Its similarity matrix also better preserves local correlations and blockwise dependency patterns, whereas the baseline exhibits larger structural discrepancies. These results support the role of $\mathcal L_{\mathrm{CFM}}$ in coordinating node and edge
evolution along the transport path. We therefore answer RQ3 affirmatively: source-target supervision improves the structural fidelity of future-graph prediction by guiding the two-stream velocity field to learn consistent node-edge dynamics.

\begin{figure*}[t]
\centering
\includegraphics[width = 0.99\linewidth]{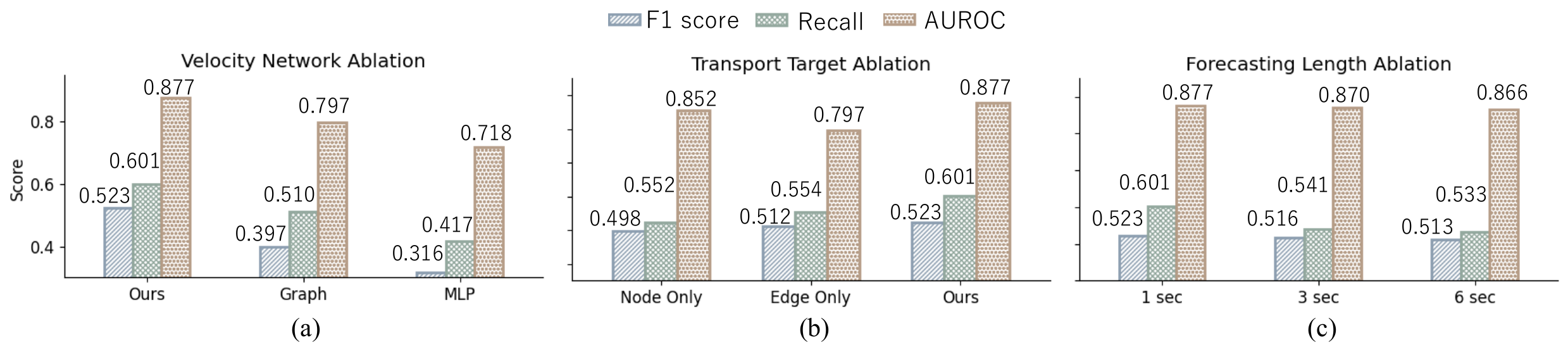}
\caption{ 
Summary of ablation study.
(a) Velocity-network parameterization,
(b) node and edge transport targets, and
(c) forecasting horizon.
The proposed joint node--edge velocity network, joint transport supervision,
and short-horizon prediction provide the strongest overall performance.
}
\label{fig:ablation}
\end{figure*}

\subsection{Ablation study}
\noindent \textbf{Effects of velocity architecture and forecasting options.}
Figure\@\ref{fig:ablation} evaluates the velocity architecture, transport
targets, and forecasting horizon. The joint velocity network achieves an F1
of $0.523$ and an AUROC of $0.877$, outperforming the graph-only
($0.397/0.797$) and MLP ($0.316/0.718$) alternatives. This result attributes
the gain to explicit node--edge coupling rather than the Flow Matching
objective alone. Joint node--edge supervision also performs best: removing
edge supervision reduces F1/Recall from $0.523/0.601$ to $0.498/0.552$,
whereas edge-only supervision retains a competitive F1 of $0.512$ but lowers
AUROC to $0.797$. Node dynamics and relational reconfiguration therefore
provide complementary predictive information. Finally, extending the horizon
from $1$\,s to $3$\,s and $6$\,s progressively decreases F1
($0.523\!\rightarrow\!0.516\!\rightarrow\!0.513$) and Recall
($0.601\!\rightarrow\!0.541\!\rightarrow\!0.533$), suggesting that the
$1$\,s target better captures rapidly evolving EEG dynamics, while longer
horizons introduce greater predictive uncertainty. 
\section{Conclusion}

In this work, we introduced \method, a temporal-conditioned graph Flow Matching framework for predictive EEG representation learning. \method decomposes node-edge sequences into past-derived states and
residuals, and learns their joint transport through a bidirectional node-edge velocity field. 
\textbf{Limitation}: Current evaluation is limited to epoched EEG and short-horizon supervision. 

\bibliography{iclr2027_conference}
\bibliographystyle{iclr2027_conference}

\clearpage
\appendix
\startcontents[appendix]
\section*{Appendix}
\label{apdx:setting}
\printcontents[appendix]{}{1}{\setcounter{tocdepth}{2}}
\clearpage


\section{Dynamic Spectral Graph Structure}\label{sec:graph}

Raw EEG signals consist of complicated neural activities that overlap in multiple frequency bands, each potentially encoding different functional neural dynamics. 
Directly analyzing EEG signals in the time domain often misses subtle state transitions that occur uniquely within specific frequency bands \citep{yang2022unsupervised,chen2023two}. 
Hence, it is beneficial to represent the intensity variations of frequency bands and waveforms by decomposing raw EEG signals into frequency components.
To effectively provide detailed insights for subtle state transitions, we perform the short-time Fourier transform (STFT) to each EEG epoch, preserving their non-negative log-spectral. Consequently, the multi-channel EEG recordings are processed as:
\begin{equation}
     \mathbf{X}_{t} = \sum_{t= \infty}^{-\infty}x[t] \,\omega[t-m]e^{-jwt},
\end{equation}
and a sequence of EEG epochs with their spectral representation is formulated as $\mathbf{X} \in \mathbb{R}^{N \times d \times T}$.

We then apply a graph representation by measuring the similarity between the spectral representation $\mathbf{X}$ across the EEG channels. Specifically, we define an adjacency matrix $\mathcal{A}_{t}(i,j)$ at each epoch $t$ as follows:
$\mathcal{A}_{t}(i,j) = \text{sim}(\mathbf{X}_{i,t},\mathbf{X}_{j,t})$
 and compute the normalized correlation between nodes $v_{i}$ and $v_{j}$, where the structure of the graph and its associated edge weight matrix $A_{i,j}$ are inferred from $X_{t}$  for each $t$-th epoch. 
 We only preserve the highest top-$\tau$ correlations to construct the evident graphs without redundancy. To avoid redundant connections and clearly represent dominant spatial structures, we retain only the top-$\tau$ strongest connections at each epoch for sparse and meaningful graph representations.
Thus, we obtain a temporal sequence of EEG spectral graphs $\{{G}_t=(\mathcal{V}_t,\mathcal{A}_t)\}_{t=0}^{T}$.

\textbf{Temporal Graph Representation.}
Taking an EEG $\mathbf{X}$ consisting of $N$ channels and $T$ time points, we represent $\mathbf{X}$ as a graph, denoted $\mathcal{G} = \{ \mathcal{V}, \mathcal{A}, \mathbf{X} \}$, where $\mathcal{V} = \{ v_1, \dots, v_N \}$ represents the set of nodes. 
Each node corresponds to an EEG channel. 
The adjacency matrix $\mathcal{A} \in \mathbb{R}^{N \times N \times T}$ encodes the connectivity between these nodes over time, with each element $a_{i,j,t}$ indicating the strength of connectivity between nodes $v_i$ and $v_j$ at the time point $t$.
Here, we redefine $T$ as a sequence of EEG segments, termed epochs, obtained using a moving window approach. 
The embedding of node $v_i$ at the $t$-th epoch is represented as $h_{i,t} \in \mathbb{R}^m$.
Specifically, we perform the short-time Fourier transform (STFT) on each EEG epoch, referring to \citep{tang2022self}.
Then we measure the similarity among the spectral representation of the EEG channels to initial the $\mathcal{A}_{t}(i,j)$ for each epoch $t$.

\section{Method and Reproducibility Details}
\label{app:conditioned-ggrf}

This appendix gives the construction and implementation details omitted from
the main paper. Throughout, \(s\) denotes physical EEG-window time, whereas
\(\tau\in[0,1]\) denotes Flow Matching transport time. The latter is an
auxiliary probability-path coordinate and is not a physiological time index.

\subsection{Prediction state and conditional reference}
\label{app:state-reference}

Each example contains twelve non-overlapping one-second windows. The first
\(H=11\) windows form the observed history
\(\mathcal C=\{(\mathbf X_s,\mathbf A_s)\}_{s=t-H+1}^{t}\), and the final
window supplies the auxiliary prediction target. Here,
\(\mathbf X_s\in\mathbb R^{N\times F}\), with \(N=19\) channels and
\(F=100\) spectral features. We construct a continuous symmetric edge state
from the cached directed graph:

\begin{equation}
\mathbf S_s
=\operatorname{offdiag}\!\left(
\frac{\mathbf A_s+\mathbf A_s^\top}{2}\right),
\qquad
\mathbf a_s=\mathcal U(\mathbf S_s)\in\mathbb R^m,
\qquad
m=\frac{N(N-1)}{2}=171 ,
\label{eq:app-edge-state}
\end{equation}

where \(\mathcal U\) extracts the strict upper triangle and
\(\mathcal S=\mathcal U^{-1}\) reconstructs a symmetric, zero-diagonal
matrix. We do not reapply top-\(k\) sparsification after symmetrization.

The node reference is the parameter-free historical mean,

\begin{equation}
\mathbf M_X(\mathcal C)
=\frac{1}{H}\sum_{s=t-H+1}^{t}\mathbf X_s .
\label{eq:app-node-reference}
\end{equation}

For the edge reference, let

\begin{equation}
\overline{\mathbf S}_{\mathcal C}
=\operatorname{offdiag}\!\left[
\frac{1}{H}\sum_{s=t-H+1}^{t}
\frac{\mathbf A_s+\mathbf A_s^\top}{2}\right],
\qquad
\overline{\mathbf a}_{\mathcal C}
=\mathcal U(\overline{\mathbf S}_{\mathcal C}) .
\label{eq:app-history-graph}
\end{equation}

A causal graph forecaster produces
\(\mathbf c_A=h_\phi(\mathcal C)\) and predicts a logit-space correction to
the historical anchor:

\begin{equation}
\mathbf m_A(\mathcal C)
=\operatorname{sigmoid}\!\left[
\operatorname{logit}\!\left(
\operatorname{clip}(\overline{\mathbf a}_{\mathcal C},
\epsilon_r,1-\epsilon_r)\right)
+\Delta_\phi(\mathbf c_A)\right],
\qquad \epsilon_r=5\times10^{-3}.
\label{eq:app-edge-reference}
\end{equation}

The forecaster uses a graph-window encoder of width \(64\), four attention
heads, one spatial layer, and dropout \(0.1\), followed by a two-layer GRU.
Because edge targets are sparse, it is trained with balanced edge MSE. For a
minibatch, define
\(\mathcal I_+=\{(b,e):|a_{t+1,b,e}|>10^{-12}\}\) and
\(\mathcal I_0=\{(b,e):|a_{t+1,b,e}|\leq10^{-12}\}\). Then

\begin{equation}
\mathcal L_{\mathrm{ref}}(\phi)
=\frac{1}{2|\mathcal I_+|}
\sum_{(b,e)\in\mathcal I_+}(m_{A,b,e}-a_{t+1,b,e})^2
+\frac{1}{2|\mathcal I_0|}
\sum_{(b,e)\in\mathcal I_0}(m_{A,b,e}-a_{t+1,b,e})^2 .
\label{eq:app-reference-loss}
\end{equation}

The forecaster is selected by full-development balanced edge MAE and is frozen
before Flow pretraining. In the headline configuration, its development
balanced MAE is \(0.202905\), compared with \(0.219361\) for persistence and
\(0.204822\) for the historical-mean reference. The terminal Flow state is
defined in residual coordinates:

\begin{equation}
\mathbf R_1^X=\mathbf X_{t+1}-\mathbf M_X,
\qquad
\mathbf r_1^A=\mathbf a_{t+1}-\mathbf m_A .
\label{eq:app-target-residuals}
\end{equation}

Thus, the node stream is history-mean-centered, whereas the edge stream is
predictor-centered.

\subsection{Conditioned graph Gaussian kernel}
\label{app:ggrf-kernel}

The source operator uses only the observed history. Let
\(d_i=\sum_j[\overline{\mathbf S}_{\mathcal C}]_{ij}\) and define

\begin{equation}
r_i=
\begin{cases}
d_i^{-1/2}, & d_i>\epsilon_D,\\
0, & d_i\leq\epsilon_D,
\end{cases}
\qquad
\mathbf R_D=\operatorname{diag}(r_1,\ldots,r_N),
\qquad
\epsilon_D=10^{-8}.
\label{eq:app-inverse-degree}
\end{equation}

The implementation constructs

\begin{equation}
\mathbf L_{\mathcal C}
=\mathbf I_N-\mathbf R_D
\overline{\mathbf S}_{\mathcal C}\mathbf R_D .
\label{eq:app-laplacian}
\end{equation}

After symmetrization, let
\(\mathbf L_{\mathcal C}=\mathbf V_{\mathcal C}
\operatorname{diag}(\lambda_1,\ldots,\lambda_N)
\mathbf V_{\mathcal C}^{\top}\). Negative eigenvalues caused by
floating-point roundoff are clamped to zero; in float32, a value below
\(-10^{-5}\) is treated as an invalid operator. The trace-normalized kernel is

\begin{align}
\widetilde\kappa_i
&=(1+\alpha\max\{\lambda_i,0\})^{-\nu},
&
\kappa_i
&=\frac{N\widetilde\kappa_i}
{\sum_{j=1}^{N}\widetilde\kappa_j},
\nonumber\\
\mathbf K_{\mathcal C}
&=\mathbf V_{\mathcal C}
\operatorname{diag}(\kappa_1,\ldots,\kappa_N)
\mathbf V_{\mathcal C}^{\top},
&
\mathbf G_{\mathcal C}
&=\mathbf K_{\mathcal C}^{1/2}.
\label{eq:app-kernel}
\end{align}

All reported full-model runs use \(\alpha=\nu=1\).
By construction, \(\operatorname{tr}(\mathbf K_{\mathcal C})=N\).

\subsection{Node source: covariance and energy}
\label{app:node-source}

For IID standard normal
\(\boldsymbol\epsilon_X\in\mathbb R^{N\times F}\), the node residual source is

\begin{equation}
\mathbf R_0^X
=\sigma_X\mathbf G_{\mathcal C}\boldsymbol\epsilon_X .
\label{eq:app-node-source}
\end{equation}

Conditioned on \(\mathcal C\),

\begin{equation}
\operatorname{Cov}(R_{0,if}^X,R_{0,jg}^X\mid\mathcal C)
=\sigma_X^2[\mathbf K_{\mathcal C}]_{ij}\mathbf 1[f=g],
\qquad
\operatorname{Cov}(\operatorname{vec}\mathbf R_0^X\mid\mathcal C)
=\sigma_X^2(\mathbf I_F\otimes\mathbf K_{\mathcal C}) .
\label{eq:app-node-covariance}
\end{equation}

Trace normalization gives

\begin{equation}
\mathbb E[\|\mathbf R_0^X\|_F^2\mid\mathcal C]
=\sigma_X^2F\operatorname{tr}(\mathbf K_{\mathcal C})
=NF\sigma_X^2 ,
\label{eq:app-node-energy}
\end{equation}

while the expected graph Dirichlet energy is

\begin{equation}
\mathbb E[
\operatorname{tr}((\mathbf R_0^X)^\top
\mathbf L_{\mathcal C}\mathbf R_0^X)\mid\mathcal C]
=\sigma_X^2F\operatorname{tr}
(\mathbf L_{\mathcal C}\mathbf K_{\mathcal C}) .
\label{eq:app-node-dirichlet}
\end{equation}

The node scale is estimated from training residuals:

\begin{equation}
\sigma_X^2
=\frac{1}{|\mathcal D_{\mathrm{tr}}|NF}
\sum_{n\in\mathcal D_{\mathrm{tr}}}
\|\mathbf X_{t+1}^{(n)}-\mathbf M_X^{(n)}\|_F^2 .
\label{eq:app-sigma-x}
\end{equation}

This gives \(\sigma_X=0.4956156682\) for the reported TUSZ experiment.

\subsection{Edge source and exact energy matching}
\label{app:edge-source}

We sample one independent innovation per unordered node pair:

\begin{equation}
\mathbf z_A\sim\mathcal N(\mathbf 0,\mathbf I_m),
\qquad
\boldsymbol\Xi_A=\mathcal S(\mathbf z_A),
\label{eq:app-edge-innovation}
\end{equation}

where
\([\boldsymbol\Xi_A]_{ij}=[\boldsymbol\Xi_A]_{ji}=z_{A,(ij)}\) for \(i<j\)
and \([\boldsymbol\Xi_A]_{ii}=0\). Define the edge-space operator

\begin{equation}
\mathbf B_{\mathcal C}\mathbf z
=\mathcal U\!\left(
\mathbf G_{\mathcal C}\mathcal S(\mathbf z)
\mathbf G_{\mathcal C}\right).
\label{eq:app-edge-operator}
\end{equation}

For output edge \(i<j\) and input edge \(a<b\),

\begin{equation}
[\mathbf B_{\mathcal C}]_{(ij),(ab)}
=G_{ia}G_{jb}+G_{ib}G_{ja}.
\label{eq:app-edge-entry}
\end{equation}

Let

\begin{equation}
\rho_{\mathcal C}=\|\mathbf B_{\mathcal C}\|_F^2,
\qquad
s_{\mathcal C}=\sqrt{\frac{m}{\rho_{\mathcal C}}}.
\label{eq:app-edge-normalizer}
\end{equation}

The implementation evaluates this quantity without materializing an
\(m\times m\) matrix:

\begin{equation}
\rho_{\mathcal C}
=\sum_{a<b}\left[
K_{aa}K_{bb}+K_{ab}^{2}
-2\sum_{i=1}^{N}G_{ia}^{2}G_{ib}^{2}
\right].
\label{eq:app-edge-normalizer-closed}
\end{equation}

The edge residual source is

\begin{equation}
\mathbf r_0^A
=\sigma_A s_{\mathcal C}\mathbf B_{\mathcal C}\mathbf z_A
=\sigma_A s_{\mathcal C}\mathcal U\!\left(
\mathbf G_{\mathcal C}\boldsymbol\Xi_A
\mathbf G_{\mathcal C}\right),
\qquad
\sigma_A=0.25 .
\label{eq:app-edge-source}
\end{equation}

Hence,

\begin{equation}
\operatorname{Cov}(\mathbf r_0^A\mid\mathcal C)
=\sigma_A^2s_{\mathcal C}^2
\mathbf B_{\mathcal C}\mathbf B_{\mathcal C}^{\top},
\qquad
\mathbb E[\|\mathbf r_0^A\|_2^2\mid\mathcal C]
=m\sigma_A^2 .
\label{eq:app-edge-covariance-energy}
\end{equation}

This matches total edge variance, not every marginal variance. Individual
edges can have different variances and nonzero cross-edge covariances. Node
and edge innovations are independent given \(\mathcal C\), but both are
shaped by the same historical graph kernel.

\subsection{Energy-matched source controls}
\label{app:source-controls}

The source study compares:

\begin{itemize}
\item \textbf{Conditioned GGRF:} every history uses its own
\(\mathbf K_{\mathcal C}\);
\item \textbf{Global GGRF:} one kernel is constructed from the average of all
training-history graphs and shared by every example;
\item \textbf{IID:} \(\mathbf K=\mathbf I_N\), yielding
\(\mathbf R_0^X=\sigma_X\boldsymbol\epsilon_X\) and
\(\mathbf r_0^A=\sigma_A\mathbf z_A\).
\end{itemize}

All arms have zero mean in residual coordinates and are centered at
\((\mathbf M_X,\mathbf m_A)\) in the original coordinates. The field receives
the per-history \((\mathbf L_{\mathcal C},\mathbf K_{\mathcal C})\) as
condition features in every arm. Thus, this comparison isolates sampled-source
covariance; it does not remove all graph conditioning from the global or IID
models.

\section{Joint Node--Edge Velocity Field}
\label{app:joint-field}

\subsection{Tokens and bidirectional interaction}

For the symmetric pair descriptor

\begin{equation}
\chi(\mathbf u,\mathbf v)
=[\mathbf u+\mathbf v;\,
|\mathbf u-\mathbf v|;\,
\mathbf u\odot\mathbf v],
\label{eq:app-pair-descriptor}
\end{equation}

the initial node and edge tokens are

\begin{align}
\mathbf h_i^{(0)}
&=P_X([\mathbf R_{\tau,i}^X;\mathbf M_{X,i}])
+P_N(\mathbf c_i),
\nonumber\\
\mathbf e_{ij}^{(0)}
&=P_A([r_{\tau,ij}^A;m_{A,ij}])
+P_P(\chi(\mathbf c_i,\mathbf c_j))
+P_O([L_{\mathcal C,ij};K_{\mathcal C,ij}]),
\qquad i<j .
\label{eq:app-initial-tokens}
\end{align}

These tokens are deterministic projections of the interpolated residual
state; the GGRF samples the residual state, not the tokens. In the headline
Flow-pretraining run, \(\mathbf c_i\) is a fixed adaptive pooling of
\(\mathbf M_{X,i}\) from 100 to 64 dimensions. In the independently trained
strict-source study, a shared channel-wise GRU additionally encodes the node
history. The global condition concatenates mean node context, edge-predictor
context, and a 32-dimensional sinusoidal embedding of \(\tau\), and modulates
each block through FiLM.

The Edge-to-Node path provides both an attention bias and an incident-edge
message:

\begin{align}
\omega_{ij}^{(\ell,r)}
&=\operatorname{softmax}_{j}\!\left[
\frac{
(\mathbf W_Q^{(\ell,r)}\widetilde{\mathbf h}_i)^\top
(\mathbf W_K^{(\ell,r)}\widetilde{\mathbf h}_j)}
{\sqrt{d_h}}
+b^{(\ell,r)}(\widetilde{\mathbf e}_{ij})
\right],
\nonumber\\
\mathbf m_i^{E\rightarrow N}
&=\frac{1}{N-1}\sum_{j\ne i}
\mathbf W_E\widetilde{\mathbf e}_{ij}.
\label{eq:app-edge-to-node}
\end{align}

After the node update, the Node-to-Edge path uses

\begin{equation}
\mathbf m_{ij}^{N\rightarrow E}
=\operatorname{MLP}_{N\rightarrow E}\!\left(
\chi(\mathbf h_i^{(\ell+1)},\mathbf h_j^{(\ell+1)})
\right).
\label{eq:app-node-to-edge}
\end{equation}

Separate node and edge feed-forward networks complete each block. The full
field uses width 64, four attention heads, two interaction blocks, dropout
zero, and separate LayerNorm-plus-linear velocity heads. The no-cross control
retains node self-attention and edge-wise processing but removes the two
cross-stream terms above.

\subsection{Conditional linear Flow Matching}

We sample one shared \(\tau\sim\mathcal U[0,1]\) and use

\begin{align}
\mathbf R_\tau^X
&=(1-\tau)\mathbf R_0^X+\tau\mathbf R_1^X,
&
\mathbf U^X
&=\mathbf R_1^X-\mathbf R_0^X,
\nonumber\\
\mathbf r_\tau^A
&=(1-\tau)\mathbf r_0^A+\tau\mathbf r_1^A,
&
\mathbf u^A
&=\mathbf r_1^A-\mathbf r_0^A .
\label{eq:app-linear-path}
\end{align}

Source and target residuals are independently paired given \(\mathcal C\).
Accordingly, this is conditional linear Flow Matching, not optimal-transport
Flow Matching. Per-coordinate velocity energies are calibrated once on the
training set:

\begin{equation}
q_X=\frac{
\sum_{n\in\mathcal D_{\mathrm{tr}}}
\|\mathbf R_1^{X,(n)}-\mathbf R_0^{X,(n)}\|_F^2}
{|\mathcal D_{\mathrm{tr}}|NF},
\qquad
q_A=\frac{
\sum_{n\in\mathcal D_{\mathrm{tr}}}
\|\mathbf r_1^{A,(n)}-\mathbf r_0^{A,(n)}\|_2^2}
{|\mathcal D_{\mathrm{tr}}|m}.
\label{eq:app-velocity-energy}
\end{equation}

The headline calibration seed is 5124, giving
\(q_X=0.4913733962\) and \(q_A=0.1262942988\). The strict source study
recalibrates the constants within each source/seed arm. The optimized objective
is

\begin{equation}
\mathcal L_{\mathrm{FM}}
=\mathbb E\!\left[
\frac{\|\mathbf v_\theta^X-\mathbf U^X\|_F^2}{NFq_X}
+\frac{\|\mathbf v_\theta^A-\mathbf u^A\|_2^2}{mq_A}
\right].
\label{eq:app-flow-loss}
\end{equation}

The implementation logs a node--edge consistency diagnostic, but its training
weight is zero in all reported experiments. Therefore, no additional
consistency objective \(\Omega\) contributes to the results.

\subsection{Transport-free downstream readout}
\label{app:no-transport-readout}

During downstream training, a channel-wise GRU augments the fixed node
condition:

\begin{equation}
\mathbf c_i^{\mathrm{down}}
=\operatorname{AdaptivePool}_{100\rightarrow64}(\mathbf M_{X,i})
+P_C\!\left[
\operatorname{GRU}(\mathbf X_{t-H+1:t,i})\right],
\label{eq:app-downstream-condition}
\end{equation}

where \(P_C\) is initialized to zero. We query the field at zero node and edge
residuals, \(\tau=1\), and physical horizon \(h=1\). The representation is the
final node hidden state after LayerNorm and before the linear velocity head:

\begin{equation}
\mathbf H_\theta^X(\mathcal C)
=\operatorname{LN}_X\!\left[
\mathbf h^{(J)}(\mathbf 0,\mathbf 0,1\mid\mathcal C)\right].
\label{eq:app-zero-query}
\end{equation}

A shared 65-parameter node head and max pooling give

\begin{equation}
\ell_i=\mathbf w^\top\operatorname{ReLU}
(\mathbf H_{\theta,i}^X)+b,
\qquad
\ell=\max_i\ell_i,
\qquad
\widehat p=\operatorname{sigmoid}(\ell).
\label{eq:app-classifier}
\end{equation}

At downstream epoch \(e\),

\begin{equation}
\mathcal L_{\mathrm{down}}^{(e)}
=\mathcal L_{\mathrm{BCE}}
+0.1\,\mathbf 1[e\geq2]\mathcal L_{\mathrm{FM}} .
\label{eq:app-downstream-loss}
\end{equation}

The field is frozen in epoch 1 and fine-tuned from epoch 2 at a smaller
learning rate. The graph forecaster remains frozen throughout.

\subsection{Datasets and Evaluation Protocols}
\label{sec:datasets}
\textbf{Datasets.}
We use Temple University Hospital EEG Seizure (TUSZ) and the TUH Abnormal EEG Corpus (TUAB) \citep{shah2018temple}, the largest publicly available EEG seizure database. TUSZ contains 5,612 EEG recordings with 3,050 annotated seizures. Each recording consists of 19 EEG channels following the 10-20 system, ensuring clinical relevance. 
A key strength of TUSZ lies in its diversity, as the dataset includes data collected over different time periods, using various equipment, and covering a wide age range of subjects.
To provide normal controls, we sample studies from the normal subset of TUAB. Unless stated otherwise, recordings are processed with the same pipeline across corpora (canonical 10–20 montage with 19 channels and unified resampling), ensuring consistent preprocessing for cross-dataset evaluation.

\textbf{Metrics.}
To answer \textbf{RQ1}, we evaluate the model using the Area Under the Receiver Operating Characteristic Curve (AUROC) and the F1 score. 
AUROC measures the ability of models across varying thresholds, while the F1 score highlights the balance between precision and recall at its optimal threshold for classification.
For \textbf{RQ2}, we measure the predicted graph structural similarity using the Global Jaccard Index (GJI) \citep{castrillo2018dynamic}:
\begin{equation}
    \mathtt{GJI}(\mathcal{E}_{true},\mathcal{E}_{Pred}) = \frac{\lvert \mathcal{E}_{true}\cap \mathcal{E}_{Pred}\rvert}{\lvert \mathcal{E}_{true}\cup \mathcal{E}_{Pred}\rvert} \quad \quad .
\end{equation}

\textbf{Model training.} 
All models are optimized using the Adam optimizer \citep{kingma2014adam} with an initial learning rate of $1 \times 10^{-3}$ in the PyTorch and PyTorch Geometric libraries on NVIDIA A6000 GPU and AMD EPYC 7302 CPU.
We adopt the adaptive Runge-Kutta NODE integration solver (RK45) with relative tolerance set to $1 \times 10^{-5}$ for training.

\subsection{Hyperparameters}
All experiments are conducted on the TUSZ and TUAB dataset using CUDA devices and a fixed random seed of 123. EEG signals are preprocessed via the Fourier transform, segmented into 12-second sequences with a 1-second step size, and represented as dynamic graphs comprising 19 nodes (EEG channels). Graph sparsification is achieved with Top-$\tau=3$ neighbors. Both dynamic and individual graphs use dual random-walk filters, whereas the combined graph employs a Laplacian filter.

\section{Additional Results}

\begin{table*}[t]
\centering
\small
\setlength{\tabcolsep}{5pt}
\caption{Ablation of pooling options on \textbf{TUSZ} (12s seizure detection) and \textbf{TUAB}. 
\textbf{Bold} indicates best result.}
\label{tab:ablation_pool}
\begin{threeparttable}
\resizebox{0.99\linewidth}{!}{
\begin{tabular}{l
S[table-format=1.3(2)] S[table-format=1.3(2)] S[table-format=1.3(2)] S[table-format=1.3(2)]
S[table-format=1.3(2)] S[table-format=1.3(2)]}
\toprule
\multirow{2}{*}{Method} 
& \multicolumn{3}{c}{\textbf{TUSZ}} & \multicolumn{3}{c}{\textbf{TUAB}} \\
\cmidrule(lr){2-4}\cmidrule(lr){5-7}
& {Acc} & {F1} & {AUROC}
& {Acc} & {F1} & {AUROC} \\
\midrule
Max pooling& $\best{0.878 \pm 0.001} $  & $\best{0.523 \pm 0.014} $   & $\best{0.877 \pm 0.003}$  &  $\best{0.787 \pm 0.004 }$ & $\best{0.786 \pm 0.004 }$  & $\best{0.877 \pm 0.006 }$\\

Mean pooling  &  0.851(3) & 0.461(2)  & 0.857(2) & 0.788(5) & 0.784(4) & 0.872(6) \\

Sum pooling &  0.867(2) & 0.506(5)  & 0.863(4) & 0.788(7) & 0.785(6) & 0.877(3) \\

\bottomrule
\end{tabular}
}

\end{threeparttable}
\end{table*}

\begin{figure}[!ht]
\centering
\includegraphics[width = 0.9\linewidth]{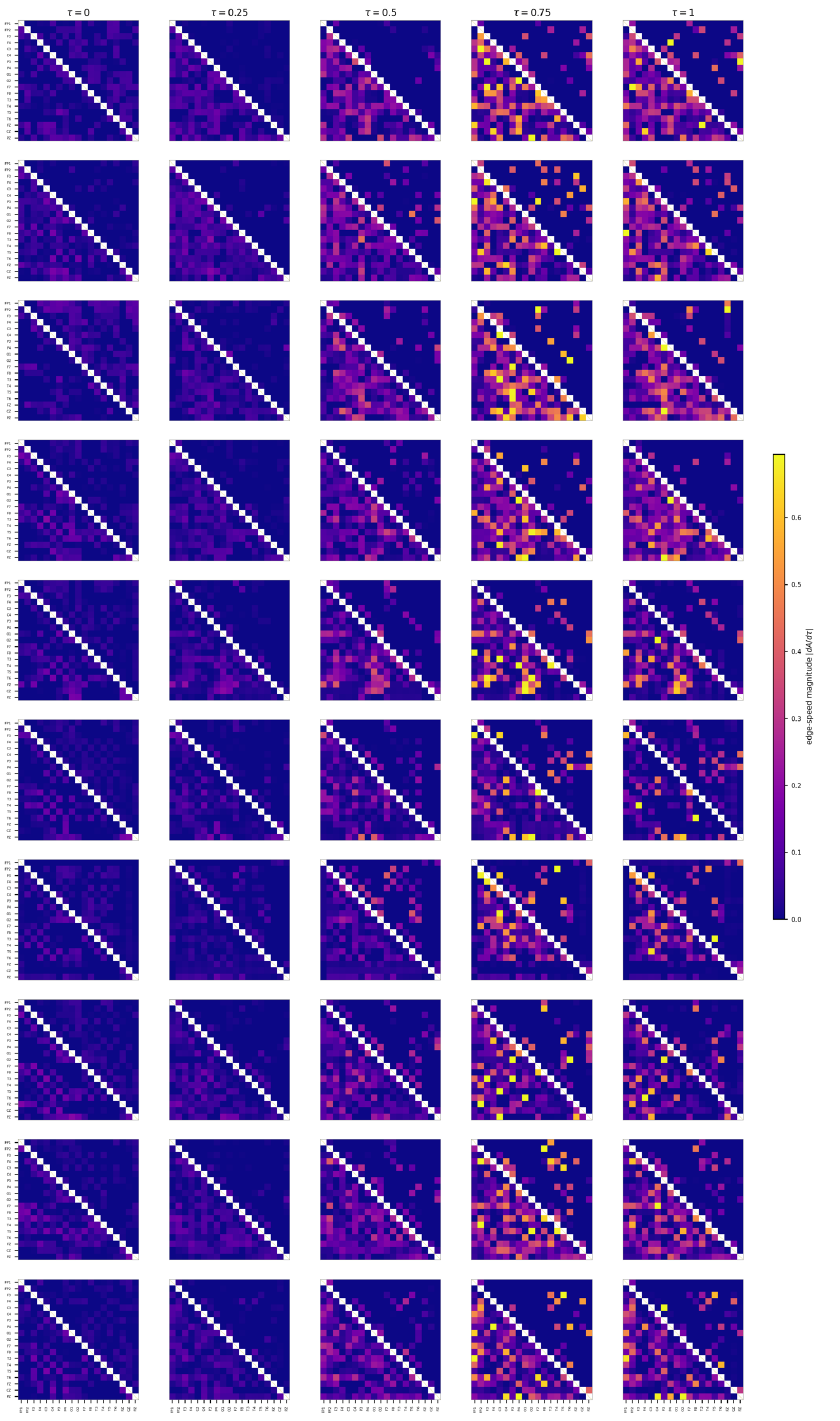}
\caption{Visualization results of latent joint node-edge dynamic field.}
\label{fig:DF}
\end{figure}

Figure~\ref{fig:DF} visualizes the learned joint node-edge velocity field
$v_\theta$ along the transport coordinate $\tau$. The field exhibits
class-dependent graph reconfiguration: seizure samples concentrate velocity
updates on specific channel pairs, whereas non-seizure samples show weaker and
sparser changes. Through bidirectional edge-to-node and node-to-edge message
passing, the field couples channel-wise residual variation with inter-channel
edge reorganization rather than evolving the two streams independently. These
distinct transport patterns make seizure-related relational structure more
explicit in the learned representation, consistent with the improved
downstream detection performance. Since $\tau$ parameterizes probability
transport rather than physical EEG time, these patterns represent conditional
node-edge transport, not physiological evolution at arbitrary time points.

\noindent\textbf{Effect of node pooling.}
Table~\ref{tab:ablation_pool} compares graph-level pooling strategies on TUSZ and
TUAB. Max pooling performs best overall, achieving an F1/AUROC of
$0.523/0.877$ on TUSZ and $0.786/0.877$ on TUAB. Its advantage is most
pronounced on TUSZ, where it improves F1 by $6.2$ and $1.7$ percentage points
over mean and sum pooling, respectively. On TUAB, all three strategies perform
similarly: max pooling obtains the highest F1 and ties sum pooling in AUROC,
while mean and sum pooling achieve a marginally higher Accuracy of $0.788$.
These results suggest that max pooling is particularly effective for seizure
detection, consistent with its ability to preserve salient channel-level
responses that may be diluted by global aggregation. We therefore adopt max
pooling for the downstream readout.

\stopcontents[appendix]

\end{document}